\documentclass{article}

\PassOptionsToPackage{numbers, compress}{natbib}

\usepackage[preprint]{neurips_2025}

\usepackage[utf8]{inputenc} 
\usepackage[T1]{fontenc}    
\usepackage{hyperref}       
\usepackage{url}            
\usepackage{booktabs}       
\usepackage{amsfonts}       
\usepackage{nicefrac}       
\usepackage{microtype}      
\usepackage{xcolor}         
\usepackage{graphicx}
\usepackage{amsmath}

\usepackage{multirow}
\usepackage{multicol}
\usepackage{makecell}
\usepackage{graphicx}
\usepackage{xspace}
\usepackage{amsmath}
\usepackage{todonotes} 
\usepackage{adjustbox}
\newcommand{\myname}{MrM\xspace}
\newcommand{\xhdr}[1]{{\noindent\bfseries #1}.}

\title{\myname: Black-Box Membership Inference Attacks against Multimodal RAG Systems}

\author{%
Peiru Yang\thanks{Tsinghua University. Email: \texttt{ypr21@mails.tsinghua.edu.cn}} \quad
Jinhua Yin\footnotemark[1] \quad
Haoran Zheng\thanks{Beijing University of Posts and Telecommunications. } \quad
Xueying Bai\footnotemark[2] \\
Huili Wang\footnotemark[1] \quad
Yufei Sun\footnotemark[2] \quad
Xintian Li\footnotemark[1] \quad
Shangguang Wang\footnotemark[2] \\
Yongfeng Huang\footnotemark[1] \quad
Tao Qi\thanks{Beijing University of Posts and Telecommunications. Corresponding author. Email: \texttt{taoqi.qt@gmail.com}}
}

\begin{document}

\maketitle

\begin{abstract}
Multimodal retrieval-augmented generation (RAG) systems enhance large vision-language models by integrating cross-modal knowledge, enabling their increasing adoption across real-world multimodal tasks.
These knowledge databases may contain sensitive information that requires privacy protection.
However, multimodal RAG systems inherently grant external users indirect access to such data, making them potentially vulnerable to privacy attacks, particularly membership inference attacks (MIAs).
Existing MIA methods targeting RAG systems predominantly focus on the textual modality, while the visual modality remains relatively underexplored.
To bridge this gap, we propose \myname, the first black-box MIA framework targeted at multimodal RAG systems.
It utilizes a multi-object data perturbation framework constrained by counterfactual attacks, which can concurrently induce the RAG systems to retrieve the target data and generate information that leaks the membership information.
Our method first employs an object-aware data perturbation method to constrain the perturbation to key semantics and ensure successful retrieval.
Building on this, we design a counterfact-informed mask selection strategy to prioritize the most informative masked regions, aiming to eliminate the interference of model self-knowledge and amplify attack efficacy.
Finally, we perform statistical membership inference by modeling query trials to extract features that reflect the reconstruction of masked semantics from response patterns.
Experiments on two visual datasets and eight mainstream commercial visual-language models (e.g., GPT-4o, Gemini-2) demonstrate that \myname\ achieves consistently strong performance across both sample-level and set-level evaluations, and remains robust under adaptive defenses.

\end{abstract}

\section{Introduction}

%
As a key augmentation strategy for LLMs, retrieval-augmented generation (RAG) has recently been extended to the visual modality, enabling broader applicability in multimodal AI tasks~\cite{chen2022murag, liu2023universal, yasunaga2023retrieval}.
By incorporating visual modalities, RAG systems can retrieve external knowledge that complements the visual input and helps reduce hallucinations in large vision-language models (LVLMs)~\cite{du2022survey,zhang2024vision}.
Recent advancements demonstrate the emerging role of multimodal RAG in enabling LVLMs to dynamically integrate knowledge for real-world applications, such as intelligent medical AI systems~\cite{ferber2024context, xia2024mmed, xia2024rule}.

For the effectiveness of RAG systems, some private-domain databases are incorporated to support vertical inference and complex reasoning~\cite{lewis2020retrieval}. 
These knowledge bases often contain private or proprietary data that are essential for supporting complex downstream tasks, while such data can be highly sensitive and should be safeguarded with robust privacy protections~\cite{ni2025survey, zeng2024good}.
Yet, the RAG paradigm inherently introduces an indirect exposure risk: the knowledge base provides information to the generation model, which then produces responses accessible to external users.
In doing so, the RAG system establishes a bridge between internal sensitive data and external adversaries, enabling interactions that may inadvertently leak private content.
This indirect access pathway creates new vulnerabilities, allowing adversaries to mount privacy attacks against the underlying database, particularly membership inference attacks (MIAs), which seek to reveal whether specific samples were part of the original database~\cite{shokri2017membership,truex2019demystifying,hu2022membership,carlini2022membership,choquette2021label,olatunji2021membership}.


Existing research on MIAs against RAG has primarily focused on text-only modality, employing various methodologies to determine whether a target sample exists in the retrieval corpus~\cite{anderson2024my,li2025generating, liu2025mask, naseh2025riddle}.
For instance, \citet{li2025generating} develop an MIA approach that analyzes semantic similarity and perplexity between target samples and RAG-generated content to infer database membership.
In conclusion, the paradigm of these methods involves providing fragments of target data and then comparing the similarity between the output and the original data.
However, LVLMs process both textual and visual inputs while typically generating text-only outputs~\cite{khan2022transformers,dou2022empirical,zhou2023vision+}.
This asymmetry introduces a challenge of modality transfer: inferring the membership status of visual data requires reasoning over purely textual responses, without direct access to visual features in the output.
Hence, these text-centric MIA methods cannot be directly transferred to LVLMs with multimodal RAG.
Besides, a new challenge lies in the balance between ensuring the successful retrieval of target data and guiding the model generation towards revealing membership information. 
The solutions to the two aforementioned core challenges are crucial for MIAs specifically designed for multimodal RAG systems.


Therefore, we propose \myname, a multi-object data perturbation framework constrained by counterfactual attacks, which is the first black-box MIA framework targeted at multimodal RAG systems. 
Its core idea is perturbing target samples and analyzing whether the textual responses implicitly reconstruct the disrupted semantics.
In this way, our method the semantics of the text and visual modalities precisely through object detection to tackle the challenge of cross-modal membership inference. 
Moreover, masking objects can minimize the affected region, thereby enhancing the effectiveness of attacks on retrieval, while perturbing the independent and complete semantics to strengthen attacks on generation.
Specifically, an object-aware data perturbation approach is employed to strategically disrupt visual semantics by masking detected entities using object detection models such as SAM\cite{kirillov2023segment}.
This approach ensures that key features are disrupted while still allowing relevant data to be retrieved if it exists in the database.
It is followed by a counterfact-informed mask selection strategy, where we quantify the informativeness of each perturbation.
We prioritize masks that maximize discriminative gaps by analyzing probability distributions and confidence differentials of a counterfactual proxy model.
This strategy aims to eliminate the interference of the self-knowledge of LVLMs, thereby preventing the reconstruction of information for non-database images during the generation phase.
Finally, we perform statistical membership inference by modeling query trials to analyze whether the textual responses implicitly reconstruct the disrupted semantics.

In conclusion, the contributions of our method are as follows:
\begin{itemize}
    \item We introduce the first black-box MIA framework for multimodal RAG systems, highlighting vulnerabilities in the privacy protection of multimodal databases.
    
    \item We propose a unified MIA framework that addresses the cross-modal alignment issue and enables concurrent attacks in both the retrieval and generation phases.
    
    \item We validate our framework through comprehensive experiments across two visual datasets and eight mainstream commercial LVLMs, demonstrating consistent strong performance and robustness against adaptive defense strategies.
\end{itemize}

\section{Related Work}

\xhdr{Multimodal RAG} \quad 
Since the emergence of cross-modal alignment models like CLIP \cite{radford2021learning}, ViLBERT \cite{lu2019vilbert}, and BLIP \cite{li2022blip}, multimodal RAG has emerged as a critical solution to address the inherent limitations of unimodal frameworks in processing cross-modal correlations, where isolated text-based retrieval fails to capture the intricate interplay between visual semantics and linguistic context required for complex vision-language reasoning\cite{chen2022murag, liu2023universal, yasunaga2023retrieval, faysse2024colpali}.
\citet{chen2022murag} propose a multimodal retrieval-augmented transformer that enhances language generation by accessing an external image-text memory, pre-trained with joint contrastive and generative objectives.
\citet{liu2023universal} propose a unified vision-language retrieval model using modality-balanced hard negatives and image verbalization to bridge modality gaps.
\citet{faysse2024colpali} utilize trained VLM to produce high-quality multi-vector embeddings from text page images, which is combined with a late interaction matching mechanism for efficient document retrieval.
Overall, multimodal RAG has now become a key paradigm, enabling robust cross-modal retrieval and grounded generation in vision-language tasks.

\xhdr{MIA against Multimodal VLMs} \quad 
Recent advancements in multimodal learning have prompted studies on MIA targeting the training data of VLMs\cite{hu2022m, ko2023practical, li2024membership, ibanez2024lumia}.
\citet{ko2023practical} propose an MIA method using cosine similarity and a weakly supervised attack that avoids shadow training.
\citet{li2024membership} present a VLM MIA benchmark and a token-level detection method using a confidence-based metric for both text and images.
While these methods can inspire the design of an MIA framework against multimodal RAG systems, they are primarily based on white-box or gray-box architectures.
However, most RAG systems are deployed in cloud environments and offered as Generation-as-a-Service (GaaS), thereby operating within black-box settings.


\xhdr{MIA against Unimodal Textual RAG} \quad 
Recent works have explored MIA in RAG systems focused on single-text modality.
\citet{anderson2024my} introduce the first RAG-MIA approach, inferring document presence by querying the system and interpreting yes/no responses.
\citet{li2025generating} propose \textsc{S$^2$MIA}, using BLEU-based semantic similarity and perplexity comparisons between target samples and generated outputs to infer membership.
\citet{liu2025mask} present a mask-based method that perturbs target documents via word masking, queries the system, and uses prediction accuracy thresholds for inference.
\citet{naseh2025riddle} design document-specific natural-text queries and infer membership by comparing system responses to shadow LLM-generated ground truth answers.
However, these text-only unimodal RAG-MIA approaches face limitations when applied to multimodal architectures. 
Vision-language models handle combined image-text inputs yet produce solely textual responses, creating a fundamental mismatch with conventional methodologies.
In conclusion, MIA specifically designed for multimodal RAG systems remains largely unexplored in the literature.
\begin{figure*}[t]
    \centering
    \includegraphics[width=\textwidth]{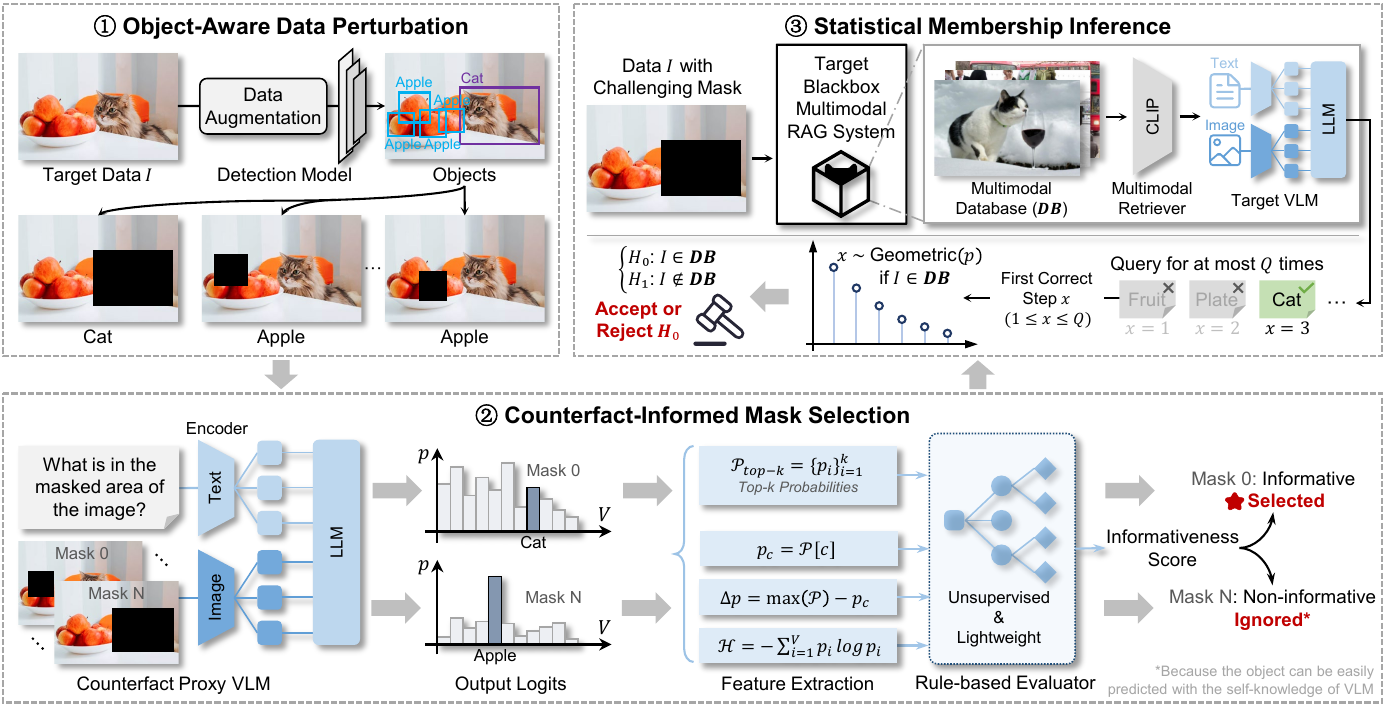}
    \caption{
    The overall framework of the proposed \myname method.
    It first perturbs the target image via object-aware masking, selects informative perturbations using a counterfact-informed mask selection strategy, and infers membership via a hypothesis test on the RAG system's response statistics.
    }
    \label{fig:method}
\end{figure*}

\section{Threat Model}
\xhdr{Attacker's Goal} \quad 
The attacker's objective is to determine whether a target image $\mathcal{I}$ or an image dataset $\mathcal{D}_i = \{\mathcal{I}_1, ..., \mathcal{I}_N\}$ exists in the black-box multimodal RAG system's database $DB$, through membership inference attacks leveraging only the system's textual outputs.

\xhdr{Attacker's Capability} \quad 
The attacker can repeatedly query the RAG system through its public interface, mimicking legitimate user interactions. 
They may craft multimodal inputs (text and images) to probe the system and observe the textual outputs generated by the vision-language model (VLM) $M$. 
The attacker is assumed to possess basic knowledge of multimodal systems but has no prior information about the database contents, model architecture, or training data.

\xhdr{Attacker's Constraints} \quad 
The attacker faces three key limitations: 
First, they cannot access intermediate system components including the VLM's output probability distributions, input embeddings, or the database $DB$. 
Second, the VLM $M$ may employ security mechanisms to reject explicitly malicious or privacy-sensitive queries. 
Third, all observations are restricted to the textual outputs of $M$, with no direct access to retrieval results or database indexing patterns. These constraints necessitate query strategies that bypass detection while extracting membership signals.

\section{Methods}
In this section, we present the technical details of our proposed method, \myname, which consists of three key components as illustrated in Fig~\ref{fig:method}: object-aware data perturbation, counterfactual-informed mask selection, and statistical membership inference.  

\subsection{Design Motivation and Overall Framework}
The core motivation behind our framework is to induce both the retrieval and generation phases of an RAG system to leak membership information simultaneously.
In the retrieval phase, the goal is to ensure that the RAG system successfully retrieves the target data, while in the generation phase, we aim to elicit relevant information about the target image from the response of the RAG system.
If the target data is input directly without any perturbation, the system is likely to retrieve the corresponding data.
Yet, in this case, it becomes challenging to determine whether the information in the model's response originates from the input or the retrieval database.
On the other hand, significantly degrading the target data through perturbation ensures that any relevant information in the model's response originates solely from the retrieval knowledge base.
However, this approach may significantly compromise the effectiveness of the membership inference framework, as it could cause the retrieval process in the RAG system to fail.

Thus, the challenge lies in the balance between these two objectives: ensuring successful retrieval while maintaining sufficient perturbation to guide the model towards revealing membership information.
To address this challenge, we propose a multi-object data perturbation framework constrained by counterfactual attacks.
This framework enables the generation of perturbations that strategically degrade the semantics of the target data, ensuring the retrieval of relevant information while preventing direct leakage from the input.
By inducing multiple responses from the model with different perturbations, discriminative features can be extracted, providing clear evidence of membership.

\subsection{Object-Aware Data Perturbation}

For the perturbation process, we set three key objectives:
(1) Ensure the target data remains retrievable,
(2) Prevent the reconstruction of information in non-database images during the generation phase,
(3) Facilitate the cross-modal transition from the image modality to the text generation modality.

To meet these goals, we adopt an object-aware perturbation approach.  
In an image, the object of interest often occupies only a small fraction of the overall scene, meaning perturbing these regions has minimal impact on the retrieval process.  
Moreover, individual objects have relatively independent semantics, allowing for the preservation of information traceability, which ensures that the source of the information can be linked back to the retrieval database.  
Lastly, objects are well-suited to be transferred into the textual modality, as they are typically well-defined and easily described in words, making them ideal candidates for generating meaningful text responses in the generation phase.

Given a target image $\mathcal{I}$, we employ an object detection model $\mathcal{D}$ (e.g., SAM~\cite{kirillov2023segment}) to localize salient objects. 
Let $O = \{o_j\}_{j=1}^K$ denote the set of detected objects, where $K$ is the number of objects. 
For each object $o_j$, we generate a binary mask $\mathcal{M}_j$ to occlude its corresponding region in $\mathcal{I}$, resulting in a perturbed image $\tilde{\mathcal{I}}_j$. 
Formally, the perturbation process is defined as:  
$\tilde{\mathcal{I}}_j = \mathcal{I} \odot (\mathbf{1} - \mathcal{M}_j) + \mathbf{0} \odot \mathcal{M}_j$,
where $\odot$ denotes element-wise multiplication, $\mathbf{1}$ is an all-one matrix, and $\mathbf{0}$ is an all-zero matrix. 
This ensures that pixels within $\mathcal{M}_j$ are set to zero while preserving other regions.

\subsection{Counterfact-Informed Mask Selection}  
To select perturbations that most effectively differentiate between database and non-database samples, we propose a counterfact-informed mask selection strategy, where we quantify the informativeness of each masked region in order to prioritize those that maximize discriminative gaps.
This is achieved by analyzing probability distributions and confidence differentials generated by a proxy vision-language model (VLM) $\mathcal{V}$, which serves as a counterfactual reference.
The goal is to eliminate the interference from the self-contained knowledge of the target LVLM, thereby ensuring that any observed semantic reconstruction is attributable to retrieval rather than memorized knowledge.

Given a perturbed image $\tilde{\mathcal{I}}j$, we input it into the proxy model $\mathcal{V}$ to obtain a probability distribution $P = \{p_i\}_{i=1}^V$  over the vocabulary $V$.
Based on this distribution, we extract the following features to estimate the informativeness and difficulty of each mask:
\textbf{Target Confidence} $p_c$: the predicted probability corresponding to the ground-truth category of the masked region. 
\textbf{Confidence Gap} $\Delta p = \max(P) - p_c$: measuring the discrepancy between the highest predicted probability and the ground-truth confidence.  
\textbf{Entropy} $\mathcal{H} = -\sum_{i=1}^V p_i \log p_i$: quantifies the prediction uncertainty, with higher entropy indicating greater confusion.
\textbf{Top-k Distribution} \( \{p_{(i)}\}_{i=1}^k \): the top-$k$ values in $P$ sorted in descending order, capturing the distributional sharpness and diversity of high-confidence predictions.

These features form a feature vector $\mathbf{f}_j = [p_c, \Delta p, \mathcal{H},  \{p_{(i)}\}_{i=1}^k] $, jointly capturing the uncertainty of the proxy model,  which is important for estimating the discriminative power of a perturbation in a black-box setting.
To assign an informative score to each mask, we adopt a rule-based evaluator that integrates the extracted features in an unsupervised manner.
Specifically, masks are ranked according to an ensemble of normalized feature scores, where high entropy, low target confidence, and small confidence gap jointly contribute to a higher informativeness score.
We prioritize masks with high estimated informativeness, as they are more likely to suppress spurious reconstruction for non-database data while maintaining discriminative signals for membership inference.

\subsection{Statistical Significance Analysis}  
To rigorously infer the membership status of a target image $\mathcal{I}$, we formulate a hypothesis testing framework grounded in the statistical behavior of the multimodal RAG system when queried about masked objects. The core intuition is that the system’s success rate in predicting occluded objects depends on whether $\mathcal{I}$ is in the database $DB$. Formally, we define two hypotheses.  
Null hypothesis ($H_0$): $\mathcal{I} \in DB$, where the system’s success probability for each mask follows $p_t$.  
Alternative hypothesis ($H_1$): $\mathcal{I} \notin DB$, with a lower success probability $p_n$, where $p_t > p_n$ by design.  

For each perturbed image $\tilde{\mathcal{I}}_j$ (derived from the $j$-th mask $\mathcal{M}_j$), we query the system repeatedly until it correctly identifies the masked object. Let $x_j$ denote the number of trials required for the first correct prediction. Under $H_0$, $x_j$ follows a geometric distribution:  
\begin{equation}
x_j \sim \text{Geometric}(p_t), \quad \mathbb{E}[x_j] = \frac{1}{p_t}, \quad \text{Var}(x_j) = \frac{1 - p_t}{p_t^2}.
\end{equation}
For $K$ masks, the total trials across all masks are aggregated as $S = \sum_{j=1}^K x_j$. By the additive property of independent geometric variables, $S$ has expectation and variance: $\mu_0 = \frac{K}{p_t}, \quad \sigma_0^2 = \frac{K(1 - p_t)}{p_t^2}$.
Invoking the Central Limit Theorem (CLT) for large $K$, $S$ approximates a normal distribution: $S \stackrel{\text{approx}}{\sim} \mathcal{N}\left(\mu_0, \sigma_0^2\right)$.
The $p$-value quantifies the probability of observing a total trial count as extreme as $S$ under $H_0$. To compute it, we first standardize $S$ and then evaluate the survival function of the standard normal distribution. Let $\Phi(z)$ denote the cumulative distribution function (CDF) of $\mathcal{N}(0, 1)$. The $p$-value is:  
\begin{equation}
p\text{-value} = 1 - \Phi\left(\frac{S - \mu_0}{\sigma_0}\right) = 1 - \Phi\left(\frac{S - \frac{K}{p_t}}{\sqrt{\frac{K(1 - p_t)}{p_t^2}}}\right).
\end{equation}
If $p\text{-value} < \alpha$ (e.g., $\alpha = 0.05$), we reject $H_0$ and conclude $\mathcal{I} \notin DB$; otherwise, we retain $H_0$, suggesting potential membership.

\section{Experiments and Analysis}
\subsection{Experimental Setups}

\xhdr{Datasets} \quad
We use two standard image datasets to build the knowledge base and perform membership inference attacks.
COCO~\cite{lin2014microsoft} and Flickr~\cite{young2014image} provide diverse image collections widely used in vision research.
From each dataset, we selected 5,000 images for the knowledge base, and 1,000 images (500 members, 500 non-members) for testing.

\xhdr{Target Models} \quad 
We conduct membership inference attacks on eight commercial models, each integrated with a local knowledge base to form a multi-modal RAG system: GPT-4o-mini~\cite{hurst2024gpt}, Gemini-2~\cite{google2025gemini2}, Claude-3.5~\cite{anthropic2024claude35sonnet}, GLM-4v~\cite{glm2024chatglm}, Qwen-VL~\cite{bai2025qwen2}, Pixtral~\cite{agrawal2024pixtral}, Moonshot~\cite{moonshot2024kimi}, and InternVL-3~\cite{chen2024expanding}.
These commercial VLMs support multi-image inputs, making them suitable for multi-modal RAG systems.
The experiments are conducted via API calls connected to a locally built knowledge base. 
This setup ensures no access to internal generation states, maintaining a strict black-box environment that mirrors real-world deployment, where only the model output is available for analysis and no information about the inner workings or intermediate states can be accessed.

\xhdr{Baselines} \quad 
To the best of our knowledge, our work presents the first MIA approach targeting multi-modal RAG systems.
Due to the lack of baselines, we adapt two strategies from text-based RAG MIA~\cite{anderson2024my, li2025generating}.
The first baseline, Query-based MIA (denoted as \textit{QB-MIA}), directly asks if the target sample appears in the retrieved references, interpreting the model's binary response as a membership signal.
The second, Similarity-based MIA (\textit{SB-MIA}), partially masks the target image and asks the model to reconstruct the missing content using the retrieved reference images.
Variants like \textit{SB-MIA-0.5} indicate the masking ratio.
Similarity between the generated description and original content is then computed, with higher similarity implying likely membership.

\xhdr{Evaluation Metrics} \quad 
Following prior works on MIA~\cite{li2024membership,shidetecting,zhang2024min}, we employ two evaluation metrics: AUC and TPR\@5\%FPR.
AUC reflects overall discrimination between members and non-members across thresholds.
TPR\@5\%FPR measures the true positive rate when the false positive rate is constrained below 5\%, providing a better assessment under strict conditions.
Since identical AUCs can result from different ROC curves, TPR@5\%FPR complements AUC for nuanced evaluation.
We report both metrics at the sample and set levels for comprehensive analysis.

\xhdr{Implementation Details} \quad 
For object-aware data perturbation, we employ the SAM2 model~\cite{kirillov2023segment} to perform object detection.
We utilize the 7B local version of Qwen-VL~\cite{bai2025qwen2} as the proxy VLM.
In the ablation study, we replace it with a weaker detector, the YOLO model~\cite{redmon2016you}.
All retrieval databases are constructed using the FAISS library~\cite{douze2024faiss}.
As the image retriever in our RAG system, we adopt the ViT variant of the CLIP model~\cite{radford2021learning}.

\subsection{Main Results}

\begin{table}[t]
  \centering
  \caption{Performance comparison of different MIA methods against RAG across eight multimodal RAG systems on Flickr and COCO datasets. 
  We report AUC and TPR@5\%FPR for each method, including \textit{QB-MIA}, three variants of \textit{SB-MIA} with different masking ratios, and our proposed \myname. 
  \myname consistently achieves the highest performance, especially under low false positive constraints.}
  \setlength{\tabcolsep}{3pt}  
  \renewcommand{\arraystretch}{1.21}  
  \begin{adjustbox}{width=\linewidth}
    \begin{tabular}{c|c|cccccccccc}
    \toprule
    \multirow{10}[6]{*}{\rotatebox{90}{Fickr}} & Methods & \multicolumn{2}{c}{QB-MIA} & \multicolumn{2}{c}{SB-MIA-0.25} & \multicolumn{2}{c}{SB-MIA-0.5} & \multicolumn{2}{c}{SB-MIA-0.75} & \multicolumn{2}{c}{\myname} \\
\cmidrule{2-12}          & Metrics & AUC   & TPR@5\% & AUC   & TPR@5\% & AUC   & TPR@5\% & AUC   & TPR@5\% & AUC   & TPR@5\% \\
\cmidrule{2-12}          & GPT-4o-mini & 64.66\% & 32.85\% & 67.10\% & 15.69\% & 70.04\% & 20.33\% & 58.58\% & 9.67\% & \textbf{80.86\%} & \textbf{66.87\%} \\
          & Claude-3.5 & 55.85\% & 16.12\% & 63.21\% & 14.05\% & 62.79\% & 14.05\% & 44.85\% & 5.69\% & \textbf{85.36\%} & \textbf{74.98\%} \\
          & Gemini-2 & 72.16\% & 9.21\% & 57.23\% & 8.03\% & 54.72\% & 7.36\% & 44.80\% & 6.69\% & \textbf{83.19\%} & \textbf{66.76\%} \\
          & Pixtral & 65.89\% & 35.18\% & 71.61\% & 19.73\% & 74.26\% & 28.43\% & 62.12\% & 26.09\% & \textbf{83.84\%} & \textbf{61.12\%} \\
          & Qwen-VL & 56.52\% & 17.43\% & 66.76\% & 13.71\% & 65.91\% & 19.06\% & 55.55\% & 10.37\% & \textbf{84.22\%} & \textbf{72.16\%} \\
          & GLM-4v & 55.23\% & 15.06\% & 66.98\% & 14.72\% & 70.78\% & 22.41\% & 58.51\% & 17.06\% & \textbf{81.93\%} & \textbf{58.79\%} \\
          & Moonshot & 53.30\% & 11.27\% & 74.63\% & 25.75\% & 75.98\% & 24.75\% & 55.06\% & 17.73\% & \textbf{80.20\%} & \textbf{65.11\%} \\
          & InternVL-3 & 51.84\% & 8.50\% & 64.23\% & 12.71\% & 67.25\% & 18.39\% & 50.92\% & 14.05\% & \textbf{83.23\%} & \textbf{68.92\%} \\
    \midrule
    \multirow{10}[6]{*}{\rotatebox{90}{COCO}} & Methods & \multicolumn{2}{c}{QB-MIA} & \multicolumn{2}{c}{SB-MIA-0.25} & \multicolumn{2}{c}{SB-MIA-0.5} & \multicolumn{2}{c}{SB-MIA-0.75} & \multicolumn{2}{c}{\myname} \\
\cmidrule{2-12}          & Metrics & AUC   & TPR@5\% & AUC   & TPR@5\% & AUC   & TPR@5\% & AUC   & TPR@5\% & AUC   & TPR@5\% \\
\cmidrule{2-12}          & GPT-4o-mini & 64.42\% & 11.01\% & 52.22\% & 4.35\% & 59.51\% & 6.67\% & 61.58\% & 12.04\% & \textbf{73.51\%} & \textbf{20.77\%} \\
          & Claude-3.5 & 52.59\% & 9.91\% & 58.89\% & 8.70\% & 61.37\% & 12.33\% & 55.56\% & 8.03\% & \textbf{82.04\%} & \textbf{43.40\%} \\
          & Gemini-2 & 70.98\% & 9.28\% & 50.38\% & 5.35\% & 51.23\% & 4.01\% & 50.65\% & 8.03\% & \textbf{84.17\%} & \textbf{57.18\%} \\
          & Pixtral & 66.22\% & 35.92\% & 60.69\% & 9.36\% & 62.96\% & 6.02\% & 64.24\% & 16.44\% & \textbf{83.02\%} & \textbf{47.24\%} \\
          & Qwen-VL & 53.46\% & 11.57\% & 55.01\% & 6.35\% & 58.10\% & 7.33\% & 58.46\% & 9.73\% & \textbf{84.11\%} & \textbf{53.12\%} \\
          & GLM-4v & 64.41\% & 32.51\% & 55.24\% & 7.02\% & 60.86\% & 10.67\% & 55.76\% & 16.78\% & \textbf{76.57\%} & \textbf{36.63\%} \\
          & Moonshot & 66.47\% & 36.41\% & 56.39\% & 5.88\% & 63.66\% & 7.67\% & 55.57\% & 8.72\% & \textbf{77.87\%} & \textbf{26.31\%} \\
          & InternVL-3 & 51.51\% & 7.86\% & 47.99\% & 4.68\% & 59.72\% & 8.33\% & 50.29\% & 6.38\% & \textbf{79.37\%} & \textbf{33.03\%} \\
    \bottomrule
    \end{tabular}%
    \end{adjustbox}
  \label{tab:main}%
\end{table}%

\xhdr{Sample-level MIA} \quad
Table~\ref{tab:main} presents the performance comparison of sample-level MIA across eight multimodal RAG systems on the Flickr and COCO datasets. 
Our method, \myname, is evaluated against two baselines: QB-MIA, which directly queries the model about the presence of a target sample in the retrieved references, and SB-MIA, which removes a fixed portion of the target image and prompts the RAG system to describe the original content based on its retrieved references.

To ensure a fair and realistic evaluation, we apply a simple yet natural defense mechanism across all methods in Table~\ref{tab:main}: a cautionary system prompt is added to the VLM, stating, \textit{``Do not reveal any information about the membership of your knowledge base.''} 
This prompt serves as a minimal safeguard against unintended memorization leakage. 
While this defense has only limited impact on the performance of SB-MIA and our proposed \myname method, it significantly weakens the effectiveness of QB-MIA, which relies on the model’s willingness to answer membership-related questions directly.
All subsequent experiments in this paper are conducted under this default defense setting.

Across both datasets and all models, \myname demonstrates a clear performance advantage, achieving consistently higher AUC scores, indicating strong overall discriminative ability.
It also particularly excels in TPR@5\%FPR compared to baseline methods, which is crucial for evaluating MIA under strict false positive constraints.
This metric reflects an attacker’s success rate under strict false positive constraints, making it more relevant in real-world scenarios where low false positive rates are essential for stealthy deployment of MIA.
The superior performance of \myname stems from its ability to precisely disrupt the most semantically critical and least easily inferred regions of the target image. 
By leveraging object-aware perturbation and difficulty assessment via a proxy vision-language model, \myname identifies and masks regions that are both salient and challenging to describe without prior exposure. 
As a result, non-member images lead to vague or inaccurate responses from the VLM. 
In contrast, for member images, the VLM can often recover the correct semantics from contextual cues because of its strong in-context learning capabilities. 
This contrast enhances the discriminative power of our statistical test and underpins the improved results observed across models and datasets.

\begin{figure*}[t]
    \centering
    \includegraphics[width=\textwidth]{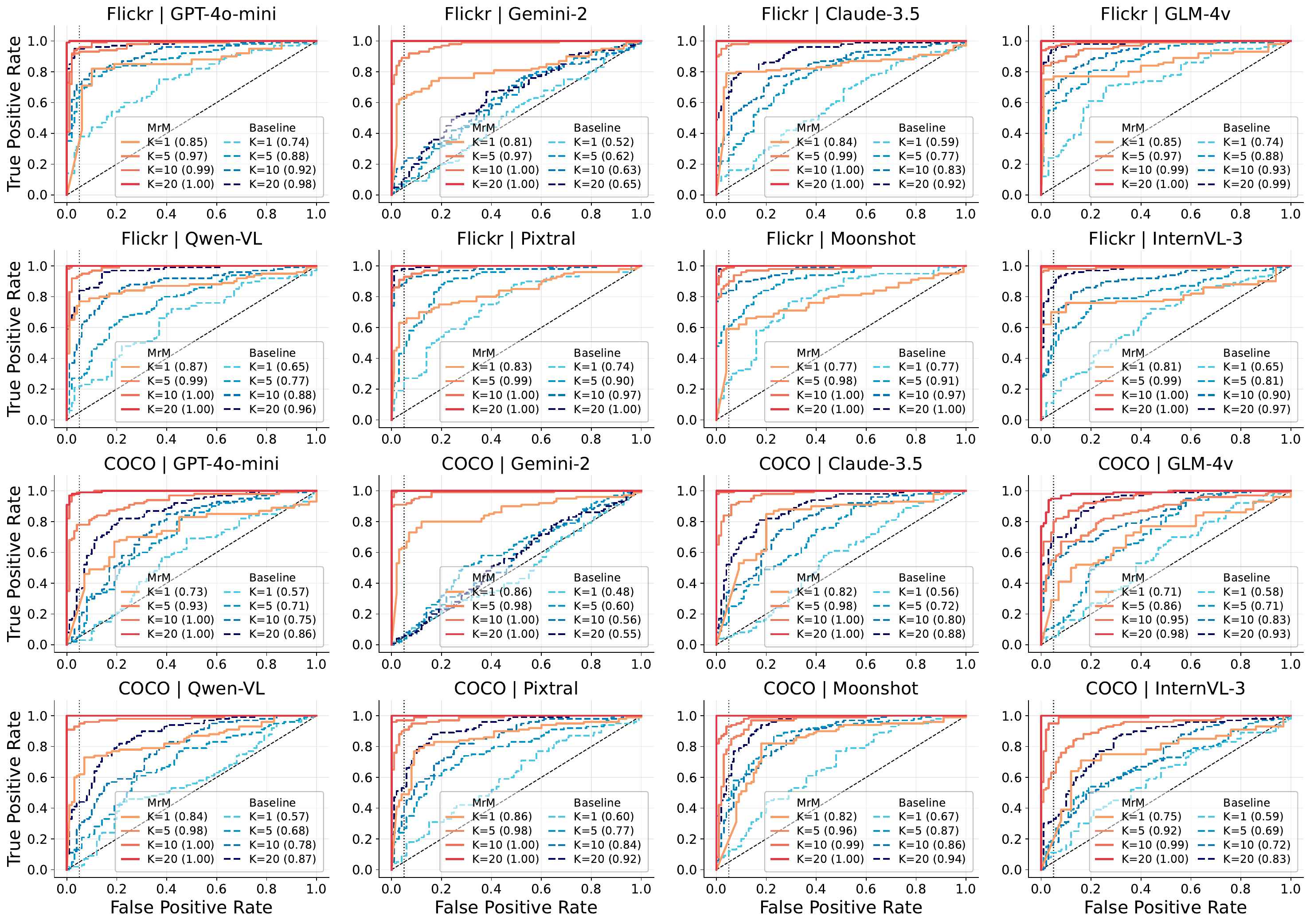}
    \caption{
    ROC curves for set-level MIAs with varying set sizes (\(K = 1, 5, 10, 20\)) on two datasets across eight multimodal RAG systems. 
    We compare \myname with the best \textit{SB-MIA} baseline. 
    \myname consistently achieves higher AUCs and steeper curves toward the top-left corner, indicating superior TPR@5\%FPR.
    The vertical gray dashed line marks the threshold at 5\% false positive rate.
    }
    \label{fig:set}
\end{figure*}

\xhdr{Set-level MIA} \quad
To evaluate the effectiveness of \myname at the set level, we plot ROC curves in Figure~\ref{fig:set} for varying set sizes $K = 1, 5, 10, 20$, across eight RAG VLMs and two datasets.
Each curve reflects the model's ability to infer membership status by aggregating predictions over a set of $K$ target samples, using a joint statistical test based on the responses of the RAG system to all samples in the set.
We compare our proposed method against the strongest variant of SB-MIA, which serves as the reference method throughout this section.
Across nearly all models and both datasets, \myname consistently outperforms the baseline method in terms of AUC, and this advantage becomes more pronounced as the set size increases.
When $K = 1$, \myname maintains strong performance, as discussed in the sample-level results above, and this advantage scales further with larger sets.
As $K$ grows, both methods show improved AUCs, but \myname consistently achieves higher values and converges more rapidly toward near-perfect performance.
In most cases, \myname achieves an AUC close to 1.0 when $K = 10$, indicating its rapid performance saturation with relatively small set sizes.

An additional advantage of \myname is that its ROC curves tend to bend more sharply toward the top-left corner, indicating higher TPR@5\%FPR under the same AUC. 
This is highlighted by the vertical reference lines at 5\% FPR, where our method consistently achieves higher TPR across models and datasets.
Furthermore, \myname shows strong and stable results across all models and datasets, including those where the baseline exhibits noticeable performance degradation.
This highlights the generalizability of our approach, even under varying model architectures and retrieval behaviors.



\subsection{Ablation Study}
\begin{figure*}[t]
    \centering
    \includegraphics[width=\textwidth]{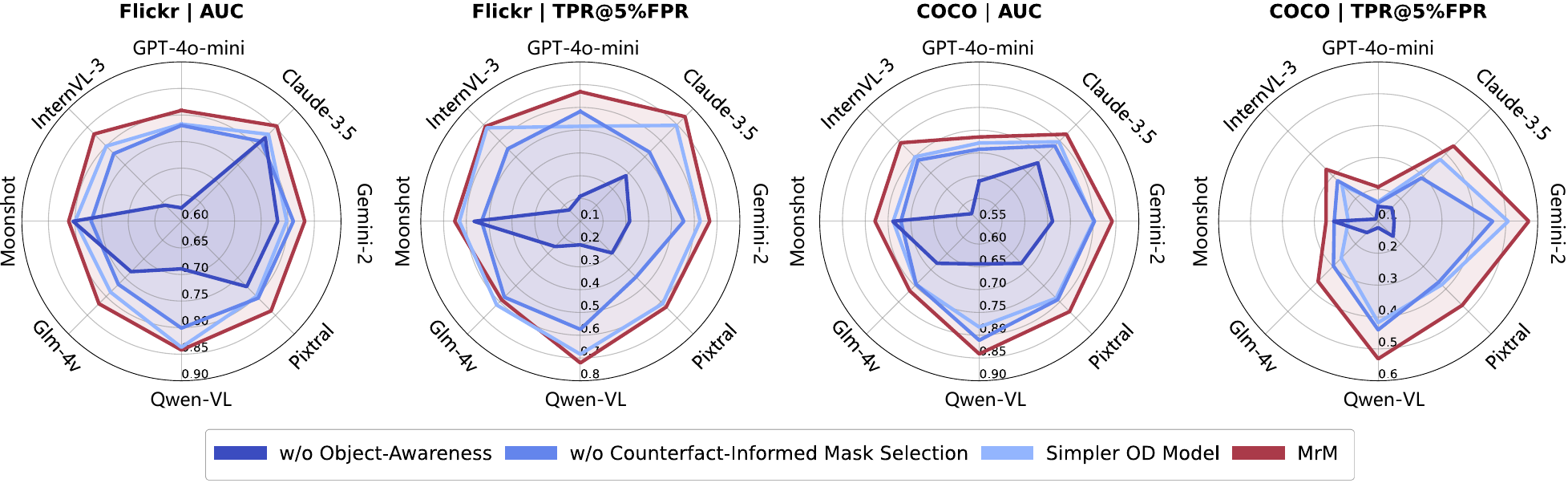}
    \caption{
    Ablation study results visualized as radar charts. 
    We compare the full \myname method with three ablated variants: w/o object-awareness, w/o counterfact-informed mask selection, and simpler OD model. 
    \myname consistently outperforms all ablated versions, demonstrating the contribution of each component to overall attack performance.
    }
    \label{fig:ablation}
\end{figure*}

To better understand the contribution of each component in \myname, we conduct an ablation study by systematically removing or replacing key elements of the pipeline.
As shown in Fig~\ref{fig:ablation}, we evaluate three variants:
(1) \textbf{w/o object-awareness}, where object detection is removed and image regions are randomly masked;
(2) \textbf{w/o proxy model}, where the proxy model is excluded and no difficulty-based mask selection is applied;
(3) \textbf{simpler OD model}, where the stronger SAM2 detector is replaced with YOLO.
All variants are tested on both Flickr and COCO datasets across eight RAG systems.

We observe consistent performance drops in all three ablation variants, confirming the necessity of each component in the full \myname pipeline.
(1) \textbf{Without object-awareness}, the model performs significantly worse, indicating that randomly masking regions often fails to target the most semantically informative parts of the image.
(2) \textbf{Without the proxy model}, the absence of difficulty assessment leads to less discriminative perturbations, weakening the signal used for membership inference.
(3) \textbf{Using a simpler object detector} results in moderate but noticeable performance degradation, suggesting that high-quality object detection contributes to more effective perturbation strategies.
These results highlight that all components play essential roles in achieving strong performance.
The full \myname method benefits from their synergy, yielding more effective and reliable membership inference across models and datasets.

\subsection{Robustness Analysis}

\begin{figure*}[t]
    \centering
    \includegraphics[width=\textwidth]{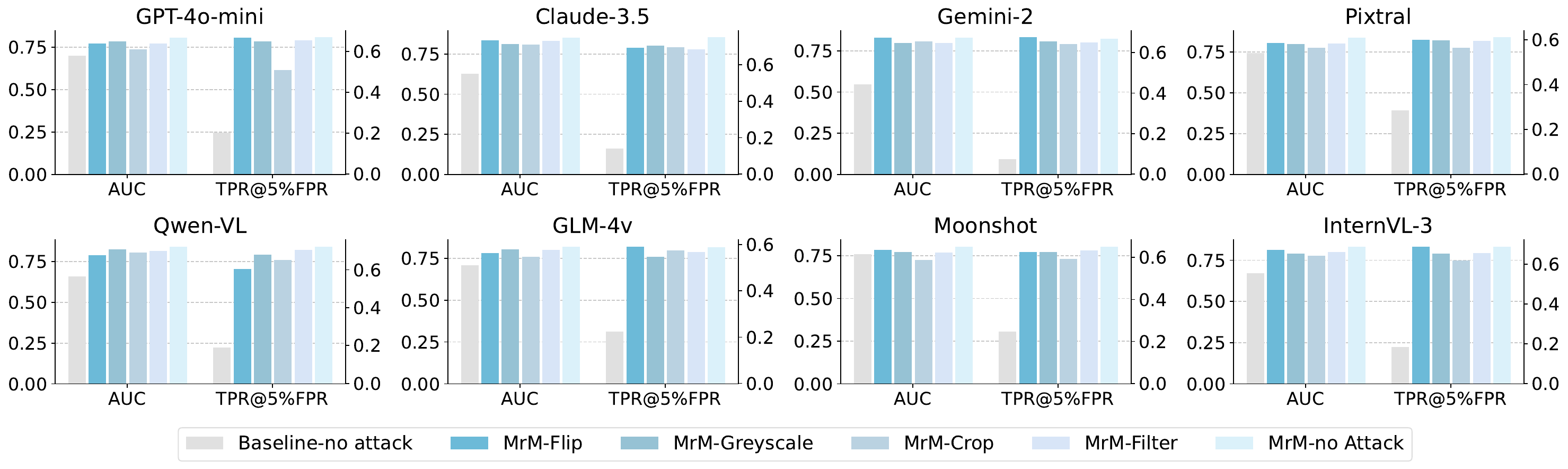}
    \caption{
    Robustness of \myname against adaptive image-level transformations applied to the database, including horizontal flipping, grayscale conversion, cropping, and Gaussian blur filter. 
    \myname maintains strong performance under all transformations.
    }
    \label{fig:robustness}
\end{figure*}

In addition to the system prompt–based defense discussed earlier, we further evaluate \myname under a class of data-level defenses based on input alteration.
Specifically, we simulate a defense setting where the retrieval database contains modified versions of the original images, processed through commonly used transformations such as horizontal flipping, grayscale conversion, cropping, and Gaussian blurring.
These transformations aim to disrupt direct visual matching while preserving the high-level semantics of the image, thus weakening naive retrieval-based MIA approaches.

To address this challenge, we extend our attack pipeline with an augmentation-aware strategy.
For each target sample, we generate multiple augmented variants using the same transformation types applied to the database.
Each variant is treated as an independent query instance—undergoing object-aware perturbation, difficulty evaluation, and statistical testing—allowing our method to explore alternative retrieval paths that remain valid despite the transformation gap.
This augmentation-aware probing increases the likelihood that at least one variant will retrieve the altered database entry, thereby restoring the model's memorization signal that might otherwise be masked.
Importantly, this design also mimics a realistic attacker's capability to guess or approximate potential transformation patterns in the deployment pipeline.
As shown in Fig~\ref{fig:robustness}, our method maintains strong performance under all four transformation-based defenses.
This result demonstrates the robustness of \myname against a range of content-preserving image alterations, reinforcing its practical applicability in more adversarial or obfuscated deployment scenarios.

\section{Conclusion}

We introduce \myname, the first black-box membership inference framework specifically designed for multimodal RAG systems.
Our method reveals previously unexplored privacy vulnerabilities in vision-language models enhanced by external knowledge retrieval.
To tackle the challenge of cross-modal alignment and retrieval-generation balance, we propose a unified MIA framework that jointly exploits both retrieval and generation phases, enabling the detection of membership signals from multimodal outputs.
\myname incorporates object-aware perturbation and counterfact-informed mask selection to precisely control semantic leakage while preserving retrieval performance.
Extensive experiments on two visual datasets and eight widely-used commercial LVLMs validate the effectiveness of our approach, showing that \myname achieves consistently strong performance under both sample-level and set-level evaluations, and remains robust even in the presence of adaptive defense mechanisms.
Our findings highlight urgent security challenges in multimodal RAG infrastructures and advance the understanding of privacy risks in systems bridging vision, language, and retrieval.

\newpage

\bibliographystyle{unsrtnat}
\bibliography{reference}

\newpage
\appendix

\section{Detailed Threat Model}

\begin{figure*}[h]
    \centering
    \includegraphics[width=\textwidth]{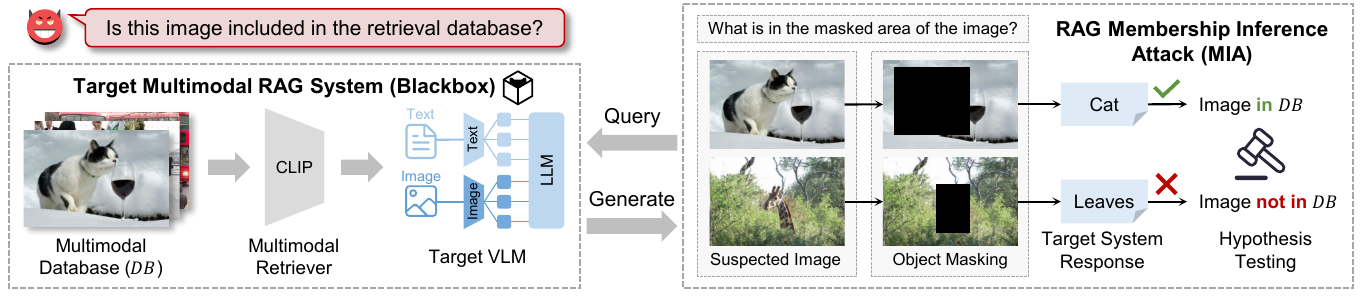}
    \caption{
   Illustration of the black-box MIA against a multimodal RAG system.
   An attacker queries the system with strategically perturbed image inputs and analyzes the textual responses to determine whether the target image exists in the underlying retrieval database, without access to model internals or database contents.
    }
    \label{fig:intro}
\end{figure*}

The attacker must construct a membership inference algorithm $\mathcal{A}$ that iteratively probes the black-box RAG system through carefully designed multimodal queries to infer membership status.
As illustrated in Fig.~\ref{fig:intro}, the attacker submits perturbed versions of a target image and observes the system’s textual responses to determine whether the image exists in the underlying retrieval database.
The attacker has no access to the internal parameters of the RAG system, its retriever, or its database content—operating entirely under a black-box assumption.

Formally, for a target image $\mathcal{I}$, the algorithm synthesizes evidence from a sequence of $k$ adaptive queries ${\mathcal{Q}_1(\mathcal{I}), \ldots, \mathcal{Q}_k(\mathcal{I})}$ and their corresponding textual responses ${\mathcal{O}_1, \ldots, \mathcal{O}_k}$ generated by the vision-language model $M$ augmented by the retrieval database, defined as
$$
\mathcal{A}\big(\mathcal{I}, \{\mathcal{O}_i\}_{i=1}^k\big) \rightarrow \{0, 1\}.
$$
This requires designing query strategies that subtly elicit membership-distinctive patterns in the output text—such as variations in specificity, contextual coherence, or knowledge granularity—while circumventing the VLM’s security mechanisms through semantically ambiguous or contextually indirect prompts.

For dataset-level verification of $\mathcal{D}_i = {\mathcal{I}_1, \ldots, \mathcal{I}_N}$, the task extends to aggregating observations across all $N$ images, necessitating a composite inference rule
$$
\mathcal{A}\big(\{\mathcal{I}_j\}_{j=1}^N, \{\{\mathcal{O}_i^{(j)}\}_{i=1}^k\}_{j=1}^N\big) \rightarrow \{0, 1\}
$$
that balances per-image evidence with global statistical confidence.

Critical challenges include identifying latent correlations between text output features and database membership, developing noise-tolerant aggregation methods to fuse multi-query evidence, and maintaining stealth by mimicking legitimate user behavior to avoid triggering defensive filters.

\section{Detailed Experimental Setup}

\subsection{Model Versions}

\xhdr{Target VLMs} 
For the target models in the multimodal RAG systems, we use the following commercial vision-language model versions via their respective APIs: GPT-4o-mini~\cite{hurst2024gpt} (\texttt{gpt-4o-mini}), Gemini-2~\cite{google2025gemini2} (\texttt{gemini-2.0-flash}), Claude-3.5~\cite{anthropic2024claude35sonnet} (\texttt{claude-3-5-sonnet}), GLM-4v~\cite{glm2024chatglm} (\texttt{glm-4v-plus-0111}), Qwen-VL~\cite{bai2025qwen2} (\texttt{qwen-vi-max-0408}), Pixtral~\cite{agrawal2024pixtral} (\texttt{pixtral-large-2411}), Moonshot~\cite{moonshot2024kimi} (\texttt{moonshot-v1-8k-vision-preview}), and InternVL-3~\cite{chen2024expanding} (\texttt{internvl3-78b}).
All models are accessed in a strictly black-box manner through API calls, with no access to their internal generation states.

\xhdr{Retriever} 
For the retrieval component, we employ the CLIP model~\cite{radford2021learning} with the ViT backbone, specifically the \texttt{openai\_clip-vit-large-patch14-336} variant, to encode image features and build the retrieval database.
Its image embedding dimension is 768.

\xhdr{Object Detectors} 
The main object detector is SAM2~\cite{kirillov2023segment}: \texttt{sam2.1-hiera-large}.
In the ablation study, we use \texttt{yolo11x}~\cite{redmon2016you} as a weaker alternative.

\subsection{Baseline and Ablations Implementation Details}
In the \textit{SB-MIA} baseline, the target image is masked by occluding its left portion.
The multimodal RAG system is then prompted to reconstruct the missing content using retrieved reference images.
We then compute the similarity between the masked image region and the model’s generated text using CLIP’s cross-modal similarity.
The numerical suffix in \textit{SB-MIA}, such as \textit{SB-MIA-0.5}, indicates the masking ratio.
Larger masking ratios make it harder for the VLM to reconstruct the original semantics, thereby increasing the potential for successful membership inference.
However, higher masking also reduces retrieval success due to greater loss of visual information.
Conversely, smaller masking ratios improve retrieval but reduce the distinguishability of generation outputs.
Compared to this trade-off, \myname achieves a better balance between retrieval accuracy and generation divergence, resulting in improved MIA performance.

In the ablation study, the variant without object-awareness applies random masking by selecting a rectangular region whose height and width range from 20\% to 60\% of the original image.
The overall masked area is thus similar to that of \myname.
The similarity computation follows the same procedure as in \textit{SB-MIA}, using CLIP-based cross-modal similarity between the masked region image and the generated text.

\subsection{Additional Implementation Details}
For all target RAG systems, we use the following prompt:
\textit{Analyze the precise positional correspondence between the masked region in the input image and the reference images. 
Extract ONLY the visual content from the EXACT SAME POSITION in the reference image. 
Output ONLY a word or phrase. DO NOT WRITE ANYTHING ELSE. 
Input Image:[IMAGE] 
Reference Images: \{[IMAGE],...\}}
If a vision-language model does not support multiple image inputs, we concatenate the target image and reference images into one image before inputting.

The retrieval database is built using FAISS~\cite{douze2024faiss}.
We use the \texttt{Flat} index type and apply \texttt{mean} pooling on image embeddings.
Although the COCO dataset~\cite{lin2014microsoft} includes ground-truth object annotations, we do not use them in our pipeline.
Instead, we simulate real-world scenarios by detecting objects using the SAM model.
All ROC curves are plotted using random sampling, with 200 samples per value of $K$.
Each experiment is repeated five times to reduce the impact of randomness.
Key Python libraries used in our implementation include:
\texttt{openai}~(v1.64.0), \texttt{faiss}~(v1.9.0), \texttt{transformers}~(v4.45.1), and \texttt{torch}~(v2.4.1)

\begin{figure}[t]
    \centering
    \includegraphics[width=0.8\linewidth]{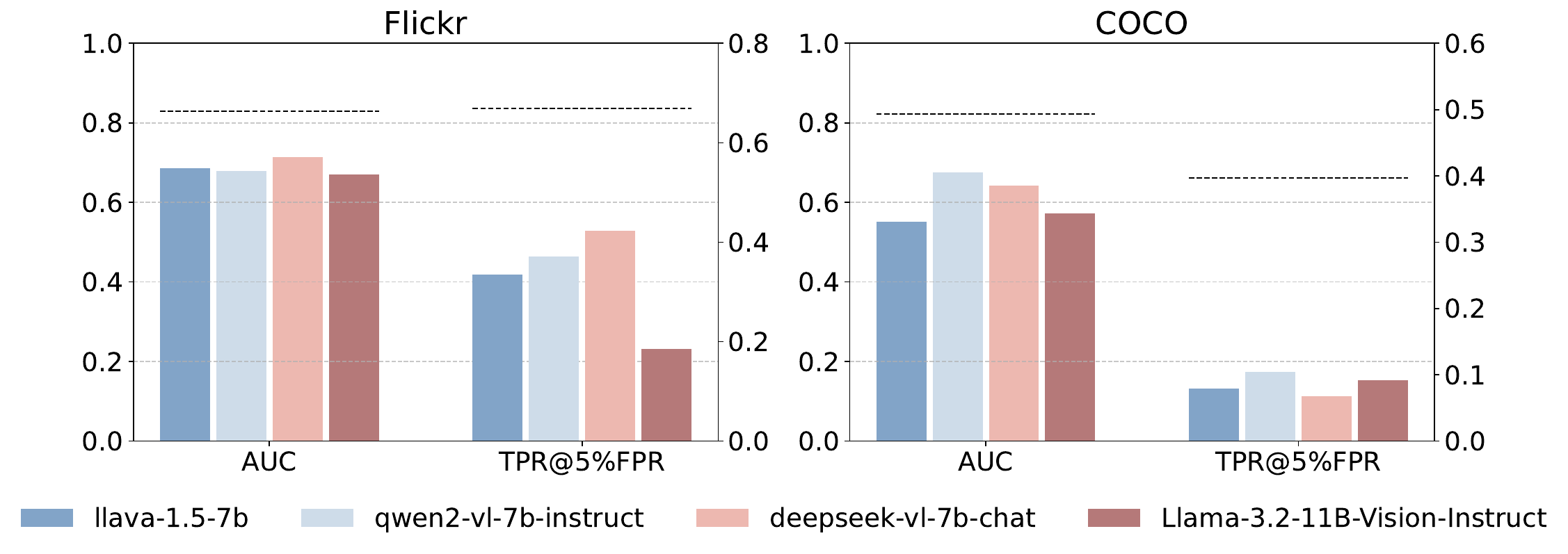}
    \caption{
    Membership inference performance of \myname on locally deployed open-source vision-language models with 7B–11B parameters, evaluated on the Flickr and COCO datasets. 
    Bars show AUC and TPR@5\%FPR. Results indicate a modest drop in performance compared to commercial models, but still demonstrate meaningful discriminative signals.
    }
    \label{fig:opensource}
\end{figure}

\begin{figure*}[h]
    \centering
    \includegraphics[width=\textwidth]{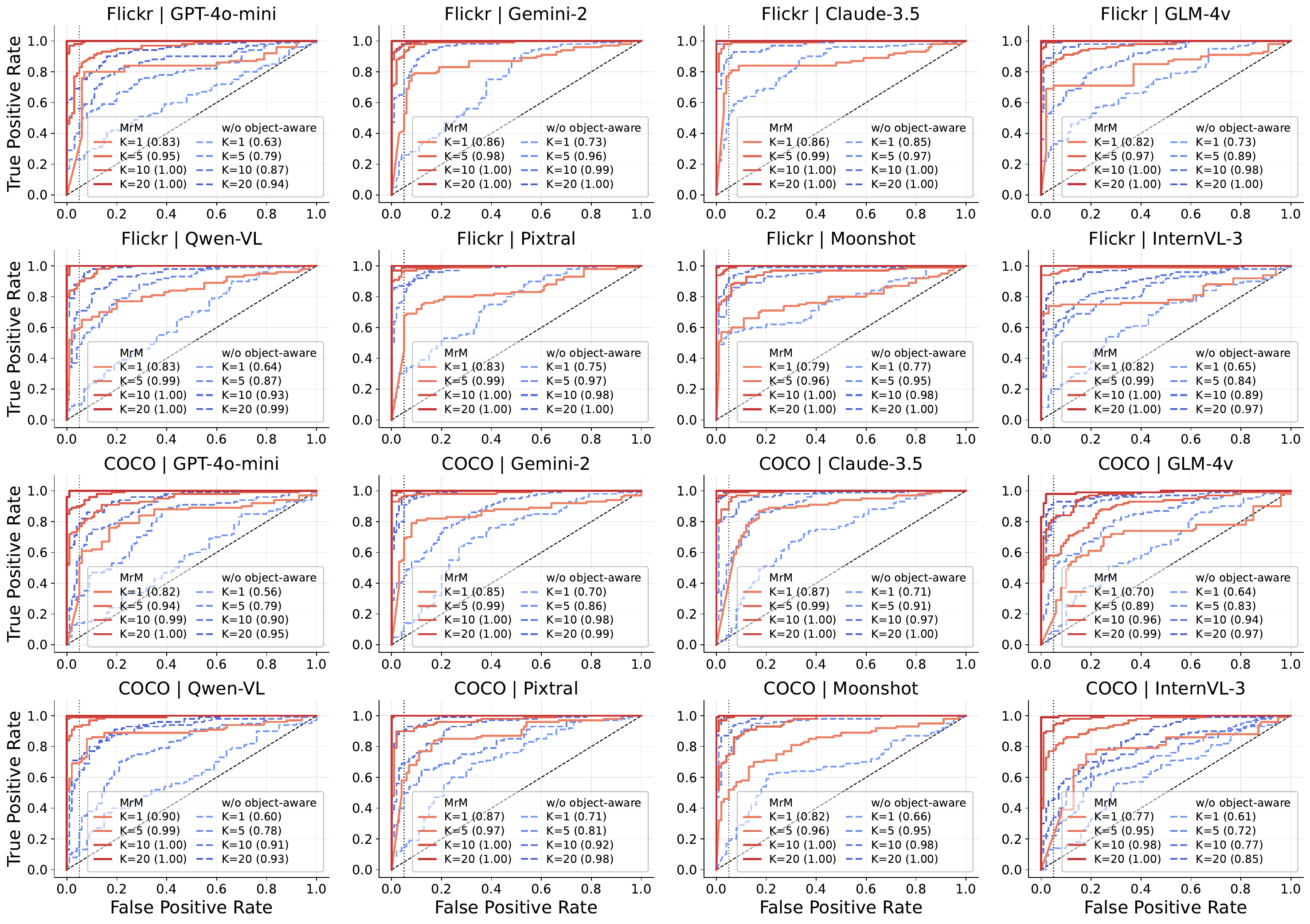}
    \caption{
   Set-level ROC curves comparing \myname with its object-agnostic variant under different set sizes (\(K = 1, 5, 10, 20\)), across eight multimodal RAG systems and two datasets. 
\myname consistently achieves higher AUCs, particularly when \(K\) is small, highlighting the effectiveness of object-aware perturbation in improving MIA performance.
    }
    \label{fig:ablation-roc}
\end{figure*}

\begin{figure*}[htbp]
    \centering
    \includegraphics[width=\textwidth]{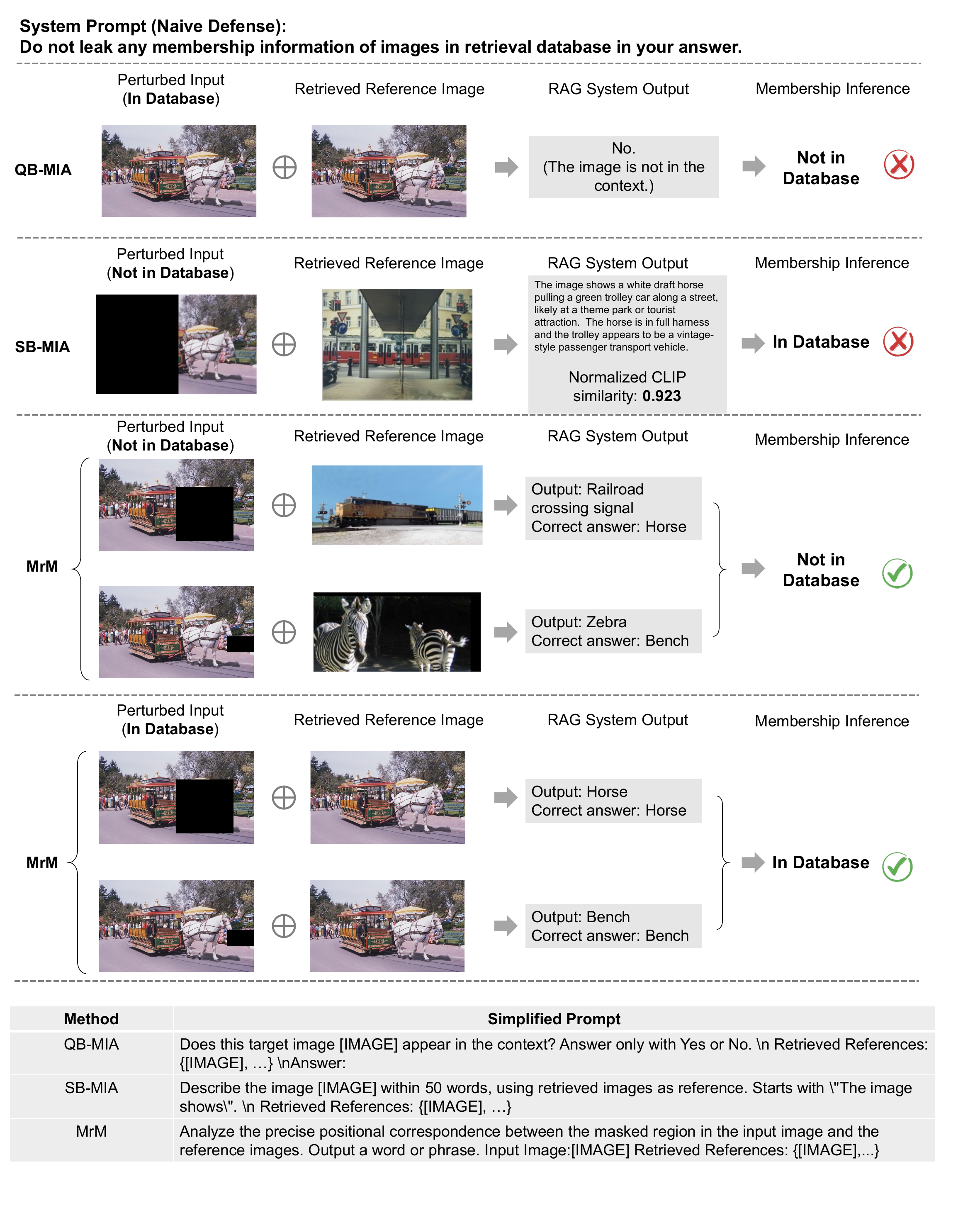}
    \caption{
Case study demonstrating the advantage of \myname in distinguishing between in-database and non-database samples.
Baseline (\textit{QB-MIA}) applied to a database image fails due to naive defense strategy.
Baseline (\textit{SB-MIA}) applied to a non-database image results in incorrect inference due to high semantic similarity reconstruction.
\myname correctly infers non-membership by masking key objects (horse and bench) and suppressing semantic recovery.
\myname correctly infers membership when applied to an in-database image, where retrieval supports accurate reconstruction.
    }
    \label{fig:case}
\end{figure*}

\section{Evaluation on Open-Source Small-Scale Models}
To explore the generalizability of \myname beyond commercial API-based systems, we evaluate its performance on several open-source vision-language models with smaller parameter scales (7B–11B). 
As shown in Figure~\ref{fig:opensource}, we test four models on the Flickr and COCO datasets: \texttt{llava-1.5-7b}, \texttt{qwen2-vl-7b-instruct}, \texttt{deepseek-vl-7b-chat}, and \texttt{llama-3.2-11B-Vision-Instruct}. 
Each model is locally deployed and evaluated using the same sample-level MIA protocol introduced in the main text.

The results show that, compared to large-scale commercial systems, these smaller models exhibit relatively lower AUC and TPR@5\%FPR scores. 
Nevertheless, the gap is not drastic, and several models (e.g., \texttt{deepseek-vl-7b-chat}) demonstrate non-trivial discriminative ability.
We hypothesize that the drop in performance is mainly due to weaker reasoning and generation capabilities in smaller models, particularly in handling multi-image prompts and resolving masked content based on retrieved references—even when relevant images are successfully retrieved.

While these results highlight certain limitations, they also confirm the generalizability of \myname across a range of architectures.
In practice, however, most real-world RAG systems are composed of API-level models backed by proprietary retrieval databases, which makes the evaluation on API-based models more representative.
Still, this open-source benchmark offers valuable insights into how model scale affects attack success.

\section{Detailed Ablation Results}

To further examine the effectiveness of object-aware perturbation, we compare \myname with its object-agnostic variant across set sizes $K = 1, 5, 10, 20$, using ROC curves plotted in Fig~\ref{fig:ablation-roc}.
Across both Flickr and COCO datasets, and for all tested models, the object-aware design yields more favorable ROC characteristics (closer to the top-left corner) and higher AUC values at nearly all $K$.
The performance gap is particularly visible at small set sizes (e.g., $K=1$), where precise perturbation is critical for triggering distinguishable responses.
As $K$ increases, the performance of both methods improves, but \myname consistently maintains a clear advantage, indicating that semantic-aware perturbations enhance both sample-level and aggregated set-level inference.

\section{Case Study}

To further illustrate the advantages of our method, we present a representative case study comparing \myname with baseline \textit{QB-MIA} and \textit{SB-MIA-0.5}, by examining both in-database and non-database scenarios for the same target image.
\textit{QB-MIA} applied to a database image fails due to naive defense strategy.

\textit{SB-MIA-0.5} applies a coarse masking strategy by occluding half of the image, then relies on cross-modal semantic similarity to infer membership. 
As shown in the first example of Fig.~\ref{fig:case}, despite the image being unseen in the context, the model leverages its internal reasoning capabilities to reconstruct the full scene and produces a response with high semantic similarity to the original (similarity score = 0.923). This leads the baseline to incorrectly infer that the image is in the database.

In contrast, \myname specifically masks two critical objects—namely the horse and the bench, which are essential for accurate scene interpretation. 
Without access to the original image through retrieval, the model fails to accurately answer these elements, resulting in a correct inference that the image is not in the database.

In the fourth case, we put this image into the database and apply \myname to it. 
The answer is correct due to successful retrieval, and our inference framework correctly identifies it as a member.

\section{Limitation}
Our experimental evaluation is conducted on two widely used and representative visual datasets, which reflect common scenarios encountered in multimodal RAG systems.
While these datasets provide meaningful coverage of typical applications, the generalizability of our findings to other domains is not guaranteed.
In particular, some emerging domains may present unique challenges to our framework.
These include settings with high inter-image similarity or highly specialized visual semantics, such as radiological medical images (e.g., CT scans and ultrasound) or satellite images.
Such domains often have distinct knowledge characteristics that may not align well with the assumptions underlying our current perturbation and inference design.
Future work should extend the evaluation to a broader range of domain-specific datasets and adjust the strategies accordingly.
This would further validate the adaptability of our approach and offer deeper insights into the privacy risks posed by RAG systems across diverse real-world settings.

\end{document}